Robotics Roadmap for Australia V.2

# Trust & Safety

Contributors: Devitt, S.K., Horne, R., Assaad, Z., Broad, E., Kurniawati, H., Cardier, B., Scott, A., Lazar, S., Gould, M., Adamson, C., Karl, C., Schrever, F., Keay, S., Tranter, K., Shellshear, E., Hunter, D., Brady, M., & Putland, T.

Version Control

| Version | Person responsible | Date Complete |
|---------|--------------------|---------------|
| V.2 | Kate Devitt | Fri 11 Sep 2020 |
| V.3 | Kate Devitt | Tue 15 Sep 2020 |
| V.4 | Maia Gould | Thu 17 Sep 2020 |
| V.5 | Kate Devitt | Fri 2 Oct 2020 |
| V.6 | Kate Devitt | Thu 5 Nov 2020 |
| V.7 | Kate Devitt | Wed 18 Nov 2020 |
| V.8 | Kate Devitt | Fri 20 Nov 2020 |
| V.9 | Kate Devitt | Fri 27 Nov 2020 |
| V.10 | Kate Devitt | Sun 14 Feb 2021 |
| V.11 | Kate Devitt | Sun 14 Feb 2021 |
| V.12 | Kate Devitt | Wed 14 Apr 2021 |

# Table of Contents

# Introduction

As robots become widespread, digitally connected, and implemented in increasingly autonomous settings, existing safety and assurances processes will be challenged across industries. To continue to make progress and innovate, ensuring that these systems have been designed responsibly and robustly will be key to safeguarding trust. While not a definitive guide, this chapter explores existing and emerging frameworks, research, and methods in robotics in Australia promoting trust and safety in robotics and autonomous systems (RAS) with increasing artificial intelligence (AI), collectively refered to as 'RAS-AI'.

While automation and automation safety has been a feature across industries in Australia for decades, the scope of tasks automated systems can complete has been enabled by the growth of digital, networked technologies and artificial intelligence. This transition – from primarily human operated to increasingly *system* or machine-operated systems – has necessitated new ways of thinking. Predicting and mitigating failures must take into account not only technical failures—such as errors in sensor measurements or system decision paths—but also human errors, both in robot design, and in interacting with robots.

This chapter considers existing processes, systems, research methods and policies and those emerging required to continue to make progress in RAS-AI. To that end we examine gaps in trust and safety between what currently exists, and what may be required in the future. As robotics become increasingly semi-autonomous or autonomous, the technical issues for trust and safety magnify.

## Regulating trust and safety

**Existing regulatory domains**: Just as new assurance methods are needed for varied technological developments, reform of how regulators operate will also be needed. The law requires that risk be minimised by engineering first, before reliance on administrative controls, i.e. human behaviour. Workplace health and safety regulations in Australia require that risk controls "so far as is reasonably practicable" be applied. This directs technologists to standards for methods of achieving safety by design.

**Robotic standards and regulation:** We may need more than ISO standards in order to evolve rapidly enough to manage assurance of robotic autonomous systems.

**Agile and Adaptive Regulation**: We need to approach regulation differently if we are to successfully adapt current regulation to autonomous systems. Regulation will never keep pace with technology; therefore regulation and codes of practice refer to use of latest standards for guidance to achieve safety, so far as is reasonably practicable.

## Human values

**Designing democratic AI:** we need to find a way to represent human values in AI – can we embed ethics in code.

**Using systems approaches to design robots:** exploring designing for trust through trans-disciplinary systems methods incorporating humanities and STEM (STEAM)[1].

---

*Autonomous, automated, AI-enabled: what is in a name?*

As the breadth and application of semi-autonomous and autonomous systems continues to expand, the definitions of these terms evolve. There is no general consensus on these terms across disciplines. There are a few reasons for this: one being differing industry needs and approaches to describing and regulating systems nationally and internationally; the other being that some of these terms have histories and applications that predate robotics.

In 2018, the US Society of Automotive Engineers (SAE) Standard J-3016 addressed this complexity directly in defining a taxonomy and levels of automation for self-driving vehicles. The SAE observed that "autonomous" a term long used in robotics to describe systems with the ability and authority to make decisions independently, had broadened to encompass entire system functionality, becoming synonymous with "automated" (section 7.1). In doing so, the term "autonomous" came to obscure dependencies systems can, and do, have on communication and cooperation with other entities, e.g. for network connectivity or data collection, or maintenance. Recognising that outside of robotics, the term "autonomy" also referred to a capacity for self-governance, the SAE observed that automated driving technologies fell short of that definition of autonomy, even when operating at level 5 – full driving automation – they still operated based on algorithms and subject to the commands of users (e.g. to go to a destination). As such, the SAE elected not to use the term "autonomous" in their taxonomy, preferring to describe five defined levels of automation.

Notwithstanding the complexity of the term, there are continuing efforts to establish a framework for levels of robot autonomy in human-robot interaction.[2] In this chapter, we recognise the popular usage of the term "autonomous" to **describe systems operating independently or semi-independently in conditions of significant uncertainty**. Automation refers to the execution of a pre-defined task within an environment of high certainty.

For brevity, throughout this chapter, the terms 'autonomous robots' or robotics, autonomous systems and artificial intelligenge 'RAS-AI' will be used interchangably to refer to both semi and fully autonomous robots. We define system assurance as providing justified confidence that increasingly autonomous systems will perform reliably and robustly, which includes safe operations.

---

[1] Bernstein, J. H. (2015). Transdisciplinarity: A review of its origins, development, and current issues

[2] https://www.ncbi.nlm.nih.gov/pmc/articles/PMC5656240/ .

## Safety and assurance

There is a long history of safety in the robotics industry. The landscape is continually changing, as new technologies emerge, societal expectations shift and change, and incidents happen. In some heavily regulated industries, safety is highly sophisticated; such as industrial robotics in car manufacturing. In emerging industries, where safety is physical and digital—there is under-investment in systems safety.

There are a number of well-developed ISO and IEC standards for safety related control systems for machines and machine specific standards including robots, AGV's and other vehicles—see Appendix A: Standard and/or project under the direct responsibility of the ISO/TC 299 Secretariat (Standardization in the field of robotics, excluding toys and military applications). For the industrial robot sector (ISO 10218 series) these have already been adopted as Australian Standards under the primary machine safety standard series AS4024.

The safety control system standards base the control system design on risk variables and lead to architecture changes in the control system as the risk increases. At the higher levels of risk this requires redundant controls with fault detection. This approach for machine safety control systems has been in place for around 25 years. International standards are also in place for service and personal care robots. Similar standards are in place and evolving for autonomous and self-driving vehicles. Current safety standards in certain industries are evolving but need investment to extend their application to new environments, use across industries, and for more than one purpose.

Robotic safety is often ensured by limiting operations to protected spaces. So, for example, drones are tested in indoor areas with mesh nets to prevent either drone or human error, but also to reduce the workload associated with bespoke regulatory compliance. Self-driving vehicles are tested with cautious collision-avoidance thresholds. The "Robot Wars" robots are encased in thick Perspex gladiatorial arenas so that human spectators are safe from shrapnel or collisions. Humans are kept apart from robots in operation wherever possible to limit the risk of harms from accidental errors—either internally driven within the robot, human error, or due to unexpected environmental constraints. Nevertheless, collaborative robots are in increasing use across industry. The application of this type of robot allows humans to work beside them without physical guarding. The term "collaborative robot" has been widely debated, since a low speed, low force robot can still cause harm dependant on the end effector. Therefore, this term is evolving to become "collaborative application".

Best practice safety management focuses on *system safety*.[3] System safety refers to risk management of an engineered system that enables balancing of safety and operability.  Systems safety applies systems engineering and systems management to identify and manage hazards in systems. Systems safety enables a systems-level risk analysis for products or services. See *Figure 1. Systems Safety Framework* and *Table 1. Defining System Safety*.

---

[3] According to NASA, "System Safety is the application of engineering and management principles, criteria and techniques to optimize safety within the constraints of operational effectiveness, time, and cost throughout all phases of the system life cycle": https://sma.nasa.gov/sma-disciplines/system-safety.

Safety Management Systems (SMSs) are the natural expansion of system safety concepts to an organisational level. ICAO describes an SMS as, 'a systematic approach to managing safety, including the necessary organisational structures, accountabilities, policies and procedures'.[4] A particular emphasis of SMSs are to imbue an organisation with a afety culture, which is 'the way safety is perceived, valued and prioritised in an organisation. It reflects the real commitment to safety at all levels in the organisation'[5]. Whilst these processes (system safety and safety management systems) are proven to drive better safety outcomes, the underlining guidelines, standards, practices associated with automation and autonomy do not exist and is an active area of research.

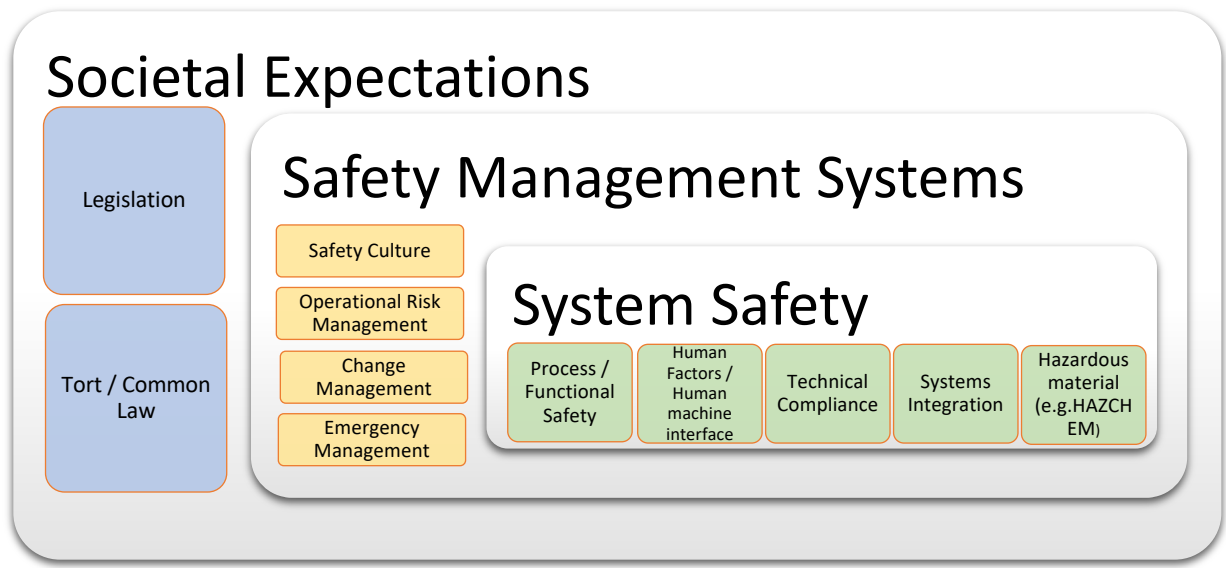

*Figure 1*. Systems Safety Framework

### Table 1. Defining System Safety

| Components of system safety | |
| --- | --- |
| **Functional safety** | The part of overall safety that depends on safety functions, the use of hardware and software or firmware, to minimize risk and keep people and assets safe (e.g., obstacle detection, collision avoidance). |
| **Human factors** | The role of people within broader systems and how the system can influence human performance. It is considered in the design of the interface between people and system, including processes, screen design and workload. |

---

[4] Annex 19 to the Convention on International Civil Aviation, 2nd Edition, ICAO 2016

[5] https://www.skybrary.aero/index.php/Safety_Culture.

| Systems of systems | How complex systems interact dynamically. A system safety approach deals with the complex behaviours resulting from their interactions. |
|---|---|
| Interactions between disciplines | Consideration of how parallel disciplines such as operational safety management, cybersecurity, resilience, change management, emergency management, and occupational health and safety interact with system safety. |
| Evolution over time | The exploration of how systems evolve over time and how to maintain safety throughout this evolution. |
| Systems integration testing | Critical to the overall safety of the system of systems, model-based testing can be applied to alleviate some of the difficulty of "in the field" testing |
| Other administrative controls | There is a host of other administrative controls that are essential to confirm safety of an automated system |

| What system and functional safety protect | The safety of the system relies on |
|---|---|
| • Safety of operator and maintainer<br>• Safety of other vehicle operators<br>• Safety of any person in the vicinity of operation<br>• Asset loss from damage to truck or collision with other vehicles<br>• Environmental spills through damage | • Safety in design<br>• Engineering controls<br>• Periodic testing of safety functions<br>• Reliability of the system<br>• Competence<br>• Well-understood failure modes<br>• Periodic testing of safety functions<br>• Clear roles and responsibilities<br>• Products and their development<br>• Ongoing maintenance and monitoring<br>• Effective change management |

## Emerging Issues in Assuring Autonomous Robots

The assurance of RAS-AI is a complex problem, increased by conditions of uncertainty.[6]

Autonomous robots differ from most machines because of the computational components that lead to their intelligence and control. They also differ from most existing software systems because of their integration with physical machines. RAS-AI rely on observations perceived by sensors to decide how they should behave and control themselves, which is then translated to actuators, enabling robots to change behaviour, or state, and change in response to their physical environments.

---

[6] It is arguable that autonomy only exists under conditions of uncertainty – otherwise it is automation: See generally, Hussein A Abbas, Darryn J Reid and Jason Scholz (Eds) Studies in Decision and Control 117 (Springer Open, 2018).

A reliable and robust autonomous robot must be able to plan and make strategic decisions about behaviour and control, despite enduring uncertainty plaguing the physical world. At the very least, these systems must be able to handle the following three types of uncertainty: Non-deterministic effects of actions, partial observability due to errors and limitations in a system's sensors and perceptions, and lack of information about the environment and its dynamics.

The enduring nature and variety of uncertainty affects all components of an autonomous robot, prompting the need to assure a system at a range of levels:

### Algorithms

Algorithms are computational methods that govern how an autonomous robot processes sensor data and makes decisions about its behaviour and control, so as to operate reliably and robustly to achieve pre-specified tasks.[7] Assurance for this component implies the need for approaches that could account for uncertainty and provide a useful guarantee of the quality of the solution proposed, and the required computational resources to compute the solution, including time and memory. State-of-the-art algorithms in autonomous systems evaluate the effects of actions prior to execution via simulation. Such approaches raise questions as to how to develop high-fidelity simulators, and how to bridge the potential discrepancy between simulators and the real world. Moreover, the high computational cost of running a high-fidelity simulator has raised questions as to whether multi-level fidelity could be used to provide the necessary quality assurance while keeping computational cost low.

### Software

It is important to differentiate between algorithms and their software implementations. Poor implementation of powerful algorithms will result in poor performance of an autonomous robot. Therefore, developers need to follow good software engineering practices. Software creation for safety related control systems of industrial robots is well described in standards such as IEC61508 and the machine specific version IEC62061. In aviation, the standard used for software design assurance onboard manned aircraft is RTCA DO-178C (and associated supplements). Furthermore, software testing should include whether the implementation of an algorithm satisfies the properties of the algorithms, both in terms of correctness and computational complexity. Examples of dangerous faults due to poor programming practices are abundant, including the unintended acceleration issue in Toyota vehicles, causing fatal crashes and a recall of multiple Toyota models.[8]

### Hardware Components.

New types of hardware, such as soft and compliant robots, pose new difficulties in assurance. Such hardware is designed to increase safety. However, the dynamics of these robots depend on the interaction between the robot and their environments, which in turn raises questions on how to properly assure such machines.

---

[7] See EASA/Daedalaen paper on Learning Assurance, which details issues associated with certain machine learning algorithms and a learning assurance process https://www.easa.europa.eu/document-library/general-publications/concepts-design-assurance-neural-networks-codann.

[8] Toyota Unintended Acceleration and the Big Bowl of "Spaghetti" Code. https://www.safetyresearch.net/Library/BarrSlides_FINAL_SCRUBBED.pdf, 7 November 2013.

Integrated Robotic System.

This assurance should ensure the system remains reliable in the event that one or more of its components are erroneous, including on seemingly simple minuscule issues, such as when the battery is low. System safety processes, such as SAE Aviation Recommended Practice 4761,[9] are being updated to focus on the complex interactions between systems, rather than at the subsystem level.

Assurance requirements pose multi-faceted issues.

1. Novel technical approaches for those assurances are required. The ability for users to easily update algorithms and software, which alters the performance of a system, implies that traditional machine assurance performed prior to deployment or sale, will no longer viable. Moreover, the high frequency of updates implies that traditional certification that requires substantial time will no longer be practical. To alleviate these difficulties, automation of assurance will likely be needed; something like 'ASsurance-as-a-Service' (ASaaS), where APIs constantly ping RAS-AI to ensure abidance with various rules, frameworks, and behavioural expectations. There are exceptions to this, such as in contested or communications denied environments, or in underground or undersea mining; and these systems need their own risk assessments and limitations imposed. Indeed, self-monitors are already operating within some systems.

2. The assurance process will require stakeholders to possess sufficient technological and computational skills. Therefore, to ensure safe operation of future robotics systems, Australia needs to educate and prepare its technology developers, certifiers, and general population for more sophisticated assurance processes.

3. What would be the suitable regulatory environment for autonomous systems? The next section deals with the existing regulatory environment for various domains and how this is evolving—or needs to evolve—with the introduction of more autonomous systems.

## Regulation in Air, Maritime and Land

The regulation of RAS-AI in Australia is quite different depending on whether a system moves on the land, flies in the air, or travels on or under the sea. Each domain has faced different levels of uptake in RAS-AI and different regulatory challenges. Arguably, the most developed is in the Air domain, with CASA facing an ever-increasing variety of flying robots. Land, with the greatest risks of human harm and most diverse rule-sets per state, is perhaps the least advanced when it comes to allowing RAS-AI on public roads—noting that the land domain does have some of the most sophisticated robots deployed in sectors such as mining in remote locations on private land and agricultural applications such as autonomous tractors, harvesters etc. Maritime has its own opportunities and challenges. There are less humans on the sea than on land, reducing the risk of RAS-AI, but there are also greater communication challenges, such as those faced by submariners.

---

[9] https://www.sae.org/standards/content/arp4761/

## Air

The Civil Aviation Safety Authority (CASA) is the Australian authority responsible for the regulation of civil aviation, including unmanned aircraft.

CASA's classification hierarchy for Unmanned Aircraft Systems (UAS) is as follows:[10]

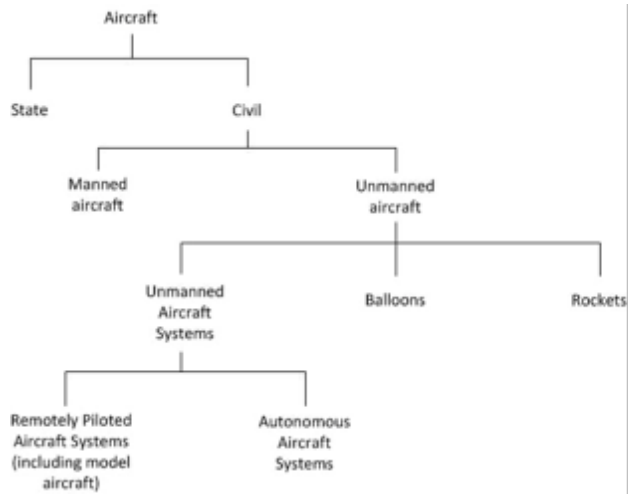

*Figure 2.* CASA Advisory Circular AC101-01 v3.0 "Remotely piloted aircraft systems – licensing and operations" published December 2019[11]

Remotely Piloted Aircraft (RPA) are included in the Civil Aviation Safety Regulation Part 101 (Unmanned Aircraft and Rockets), and are the subject of CASA guidance material. Whilst there is a regulatory provision specifying that a person cannot release an autonomous aircraft without an approval from CASA (regulation 101.097 of CASR 1998) there is no specified content for Autonomous Aircraft Systems (AAS). Current practice sees the requirements for RPA applied to all UAS, without reference to level of autonomy or correspondingly graduated requirements. Therefore, it can be implied that the existing guidance is a blanket set of standards for all RPA, ranging from remotely piloted hobbyist drones to fully autonomous aircraft flying in civil airspace.

This lack of graduated approach will pose a challenge for CASA, as the needs and requirements of an RPA being controlled by a human will differ dramatically to that of a fully autonomous aircraft. The human-in-the-loop component will influence a number of elements, including communication interface, visual capabilities, potential safety risks, intent, etc. As autonomous aviation technology continues to progress, a more robust regulatory framework will need to be developed in parallel. One benefit of the current regulatory framework is that it is extremely risk based. One of the overarching requirements to operate outside of the standard operating conditions is to ensure that there is no adverse risk to aviation safety. Guidance to ensure that the regulatory requirements are met is found within the Joint Authorities for Rulemaking on Unmanned Systems (JARUS) Specific Operations Risk Assessment (SORA) process, which CASA has adopted as a risk assessment methodology for unmanned aircraft.[12]

---

[10] Technically a rocket is not an aircraft under the definition of 'aircraft' in the Civil Aviation Act 1988 (unless it is a rocket-powered aircraft)

[11] https://www.casa.gov.au/sites/default/files/101c01.pdf.

An additional challenge will be safely and responsibly integrating autonomous aircraft with manned aircraft, such as regular passenger transport (RPT) services. At this stage, RPA usually operate in low-level segregated airspace. However, with the anticipated expansion of the industry it can be foreseen that RPA and manned aircraft will eventually need to operate within the same airspace, considering the full breadth of possible applications of autonomous aircraft, e.g. medical services, delivery services, passenger transport, etc. A common communication interface between RPA, manned aircraft and air traffic controllers will be required for safe and seamless integration of autonomous aircraft in the existing civil aviation system.

## Maritime

The Australian Maritime Safety Authority (AMSA) is the Australian authority responsible for maritime safety, protection of the marine environment from pollution, and search and rescue. As part of these responsibilities, AMSA regulates vessels operating within Australia's Exclusive Economic Zone (EEZ), including vessels capable of autonomous and remote-controlled operation.

The laws, Marine Orders, and standards that apply to all commercial vessels were written for traditional manned vessels, but remotely operated and autonomous vessels must also comply with them. As the unmanned vessels generally cannot comply with the design, construction, equipping and survey requirements applied to traditional vessels, and there are no tailored standards available to use, operators must seek exemptions in order to operate. This reliance on exemptions may not be feasible beyond the short term, due the administrative burden and delays it creates for operators and AMSA.

---

[12] https://www.casa.gov.au/publications-and-resources/media-hub/speeches-and-presentations/rpas-australian-skies-conference.

Case study: Autonomous Vessel Forum 2019

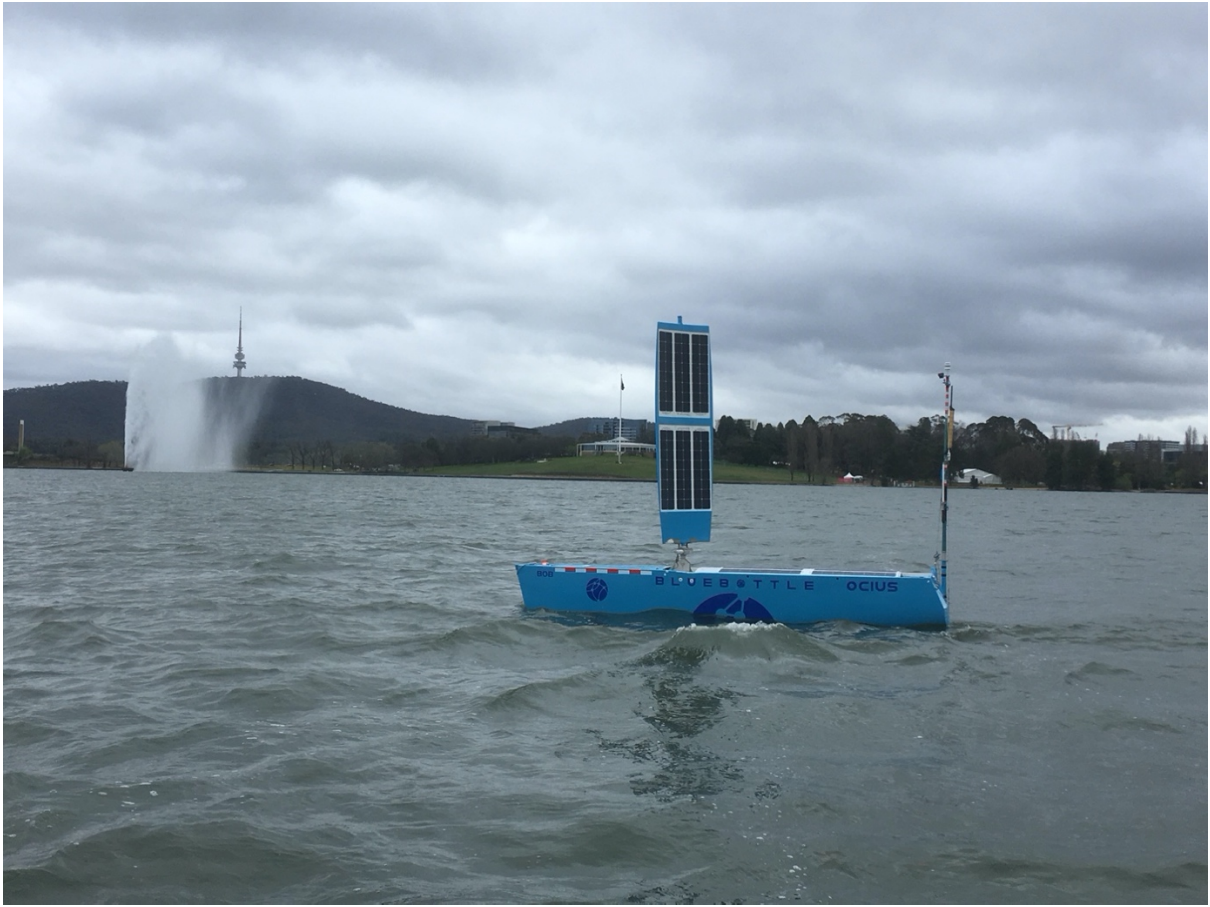

*Figure 3.* Ocius Bluebottle on Lake Burley Griffin at the Autonomous Vessel Forum 2019, with permission from Ocius.

The Autonomous Vessel Forum 2019 explored the challenges, opportunities and risks in regulating autonomous and remotely operated vessels. Hosted by the Australian Maritime Safety Authority, in partnership with TASD CRC, the Autonomous Vessel Forum 2019 saw 135 domestic and international experts from industry, academia, Navy, and Government, come together to share their knowledge and learnings in automation and digitalisation.

The Autonomous Vessel Forum 2019 marked a positive step towards regulator and industry collaboration to ensure thorough testing, trials and safe use of the systems, infrastructure and technology gaining traction in Australia's maritime industry. The forum also reinforced the vital roles that professional seafarers will continue to have in Australia's maritime industry.

The key forum themes were:

1. Success in automation and remote operation is achieved through continued **collaboration and partnership.**
2. Uptake of automation and remote operation is **driven by environmentally-friendly solutions and commercial efficiency.**

3. Proven and assured technology offers **better safety for people** and protection of Australia's marine environment.
4. **Understanding risks** and **implementing risk controls** for remotely operated and autonomous vessel technology is crucial to safe operations.
5. Building **assurance** in the **behaviours and functions of a system**, rather than specific technology, will define a 'safe system'.
6. Regulators need to **enable testing, trials and safe failure** of both technology and regulation.
7. Australian and international laws don't recognise remote and autonomous operation. However, **flexibility and objective-based solutions** allow remote and autonomous vessels to operate. There are often underutilised and present opportunities.
8. Automation and remote operation is happening today, and **seafarers are essential** to its success. Jobs at sea are changing, with new opportunities emerging alongside new technology.
9. Cyber security is an industry issue, but it's important to **regulate the need to detect, respond and recover** from a cyber-attack.

Further information is available on the [AMSA website](#).

AMSA's challenge is to adapt longstanding regulatory and operational arrangements to provide for the safe operation of unmanned vessels, in a way that will be effective in the short, medium and long term. AMSA is taking a collaborative approach and is actively engaging with leaders and key stakeholders in the fields of autonomous and remotely operated vessel design, technology, operation, and regulation. These stakeholders include TASD CRC, which has a program underway to explore assurance of autonomous systems and identify accreditation pathways. AMSA will leverage its experience and that of its stakeholders, to identify the best way to provide effective regulation, ensure the safety of people and vessels, and protect the marine environment in Australia.

Once an improved regulatory approach is implemented, the assurance and accreditation process will be streamlined, provide a more appropriate match of risk to regulatory overlay, and it will no longer be a barrier to the uptake of emerging technology in the maritime domain.

## Land

### Case study: Evolution of the Australasian New Car Assessment Program (ANCAP)

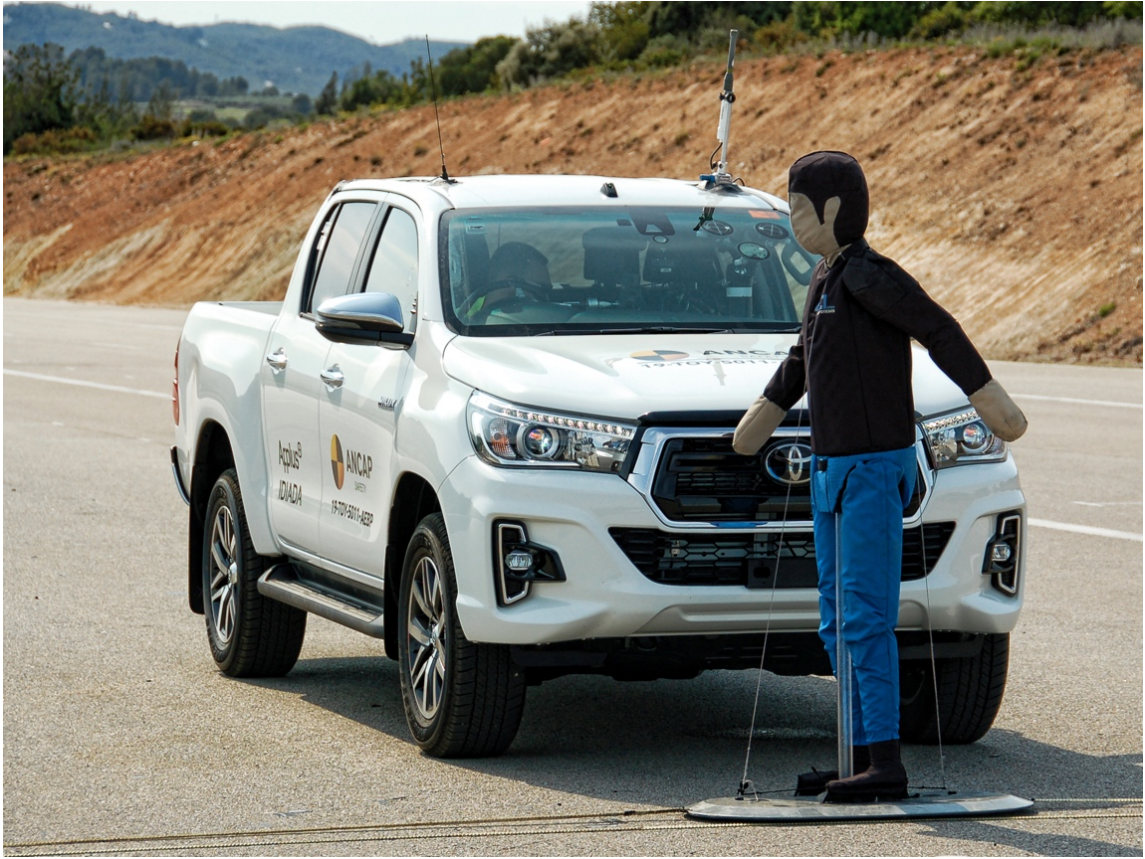

*Figure 4.* AEB (car-to-pedestrian) Toyota Hilux with permission from ANCAP

When autonomous robots become consumer products, reliability and robustness must be conveyed in an accessible manner to consumers. An example of such information is the ANCAP rating for passenger vehicles. Aside from helping consumers, such a rating increases the safety standard of the entire industry: From 2002 to 2014, the percentage of vehicle types sold in Australia with a 5-star ANCAP rating has grown from 0 to 75%[13], as car manufacturers attempt to ensure the competitiveness of their vehicles. It seems natural to bring such safety ratings to autonomous robots. However, this is not easy. Existing rating mechanisms was designed for slowly changing systems. In contrast, autonomous robots depend on algorithms and software that learn and adapt its output regularly. Work has started on developing a more suitable rating mechanism for autonomous robots, such as the Euro NCAP Driver Assistance Systems[14] and a project of the ANU in collaboration with York University in the UK, under the Assuring

---

[13] https://www.york.ac.uk/assuring-autonomy/projects/atm/.

[14] https://www.euroncap.com/en/vehicle-safety/safety-campaigns/2018-automated-driving-tests/.

Autonomy International Programme.[15] Many other issues and potential solutions are being explored, compiled, and developed as a Body of Knowledge for the Assurance of Robotics and Autonomous Systems,[16] including questions on who should perform the test and how often? A balance between convenience and reliability is necessary, which includes considering what kind of training should be provided to the evaluator and what level of expertise would be suitable? Higher requirements mean fewer qualified evaluators, increasing the difficulty for regular re-testing.

## Private land: Mining

The mining industry is highly regulated, similar to many industries, due to the significant safety risk. As the robotics industry grows, and there are more robotic applications being developed or applied, the concept of safety by design is critical to the success of industry growth. There are many lessons, experiences, and collaborations across industry sectors in the field of robotics and autonomous capability.

Australia's deployment of autonomous production drills and autonomous haulage systems (AHS) over the last 25 years in key locations like the Pilbara is world-leading in large scale adoption and use of autonomous vehicles. The relationship between the mining companies and the regulator in Western Australia is strong and there was a proactive effort by the mining companies to engage early on establishing a code of practice to be used by the industry in Western Australia.[17] This guide is used all over the world—it has even been quoted by mine operators in the Oil Sands of Canada as being "the Bible for safe operations of autonomous equipment".

Functional safety, as originally defined in IEC61508, is a key component of systems safety and has been actively adopted by the leading mining companies that have deployed autonomous systems (Autonomous Haulage Systems and Autonomous Drilling). The Global Mining Guidelines Group recently published a guideline for applying functional safety to autonomous systems in mining, which originated from the work on the Implementation Guideline for Autonomous Systems (version 1, version 2 is currently in development).[18]

This guideline outlines layers of an overall autonomous operating environment, and although this has come from a mining sector perspective, it can be applied across the board to a wide variety of robotic capabilities. There is a consideration from a product lifecycle perspective, and how this interacts with the specific operating environment – the application lifecycle.

## Public land

While the licensing and operation of motor vehicles is regulated at a state and territory level, there is Federal oversight, particularly at a broader policy and reform level. The National

---

[15] https://www.york.ac.uk/assuring-autonomy/projects/atm/.

[16] https://www.york.ac.uk/assuring-autonomy/body-of-knowledge/.

[17] Code of Practice: Safe mobile autonomous mining in Western Australia. http://www.dmp.wa.gov.au/Documents/Safety/MSH_COP_SafeMobileAutonomousMiningWA.pdf

[18] https://gmggroup.org/guidelines/.

Transport Commission (NTC) and Austroads are instrumental in the development of Australia's policy in relation to RAS-AI for land transport. However, they have advisory capacity only and cannot make regulation or enforce compliance.

Some of the key organisations involved in the policy and reform for autonomous vehicles are set out below:

**Automated Vehicle Decision Making and Priority Setting**

| Transport and Infrastructure Council |
|---|
| Makes decisions on national reforms to improve the efficiency and productivity of Australia's infrastructure and transport systems |
| Sets national reforms priorities. Current priorities include removing barriers to innovation and capitalising on new and emerging technologies |
| **Transport and Infrastructure Senior Officials' Committee** |
| Advises and assists the Transport and Infrastructure Council on all non-infrastructure priorities |

**Australian Government Automated Vehicle Roles and Responsibilities**

| Department of Infrastructure, Transport, Cities and Regional Development | National Transport Commission | State and territory transport and road agencies | Austroads |
|---|---|---|---|
| **Office of Future Transport Technology**<br>Coordination across portfolios<br>Land transport technology policy framework and action plan<br><br>**Vehicle Safety Standards Branch**<br>Importation and first supply of automated vehicles<br>Review of Australian Design Rules<br>International standards harmonisation | Develop and propose national law reform to enable the commercial deployment of automated vehicles.<br><br>Current automated vehicle reforms:<br>• In-service safety for automated vehicles<br>• Government access to vehicle generated data<br>• Motor accident injury insurance and automated vehicles | Responsibilities include:<br>• In-service vehicle regulation<br>• Vehicle registration<br>• Road rules and driver licensing<br>• Road management<br>• Approval/ regulation of automated vehicle trials | Conducts road and transport research to inform policy development and guidance on the design, construction and management of the road network and its associated infrastructure.<br>Current automated vehicle projects:<br>• Infrastructure changes to support automated vehicles on rural and metropolitan highways and freeways<br>• Pavement markings for machine vision<br>• Integrating advanced driver assistance systems in driver education |

*Figure 5*. NTC Automated Vehicle Program, National Transport Commission, October 2019[19]

In addition to the above stakeholders, the Australian Road Research Board (ARRB), an Australian/New Zealand government research and advisory organisation, is heavily involved in this field. In 2016, ARRB established the Australian New Zealand Driverless Vehicles Initiative (ADVI) which has over 180 members from industry, governments, academia, and international organisations.

The NTC, working with the various stakeholders, is leading an Automated Vehicle Program, with the goal of providing end-to-end regulation to support the safe commercial deployment and operation of automated vehicles at all levels of automation in Australia.

The current taxonomy for levels of automation is based on the Society of Automotive Engineers (SAE) International Standard J3016 as follows:

---

[19]https://www.ntc.gov.au/sites/default/files/assets/files/NTC%20Automated%20Vehicle%20Reform%20Program%20Approach%20%28October%202019%29%20-%20Public%20version.pdf.

| Levels of vehicle automation | | | | | | |
|---|---|---|---|---|---|---|
| | **Level 0** | **Level 1** | **Level 2** | **Level 3** | **Level 4** | **Level 5** |
| **Vehicle's role** | Nothing | Accelerates and brakes OR steers e.g. cruise control | Accelerates and brakes AND steers e.g. automated reverse parking | Everything, only under certain conditions eg. specific locations, speed, weather, time of day | Everything, only under certain conditions e.g. specific locations, speed, weather, time of day | Everything |
| **Human driver's role** | Everything | Everything but with some assistance | Remains in control, monitors and reacts to the driving environment | Must be capable of regaining control on request when vehicle is driving | Nothing when vehicle is driving, but everything at other times | Nothing |

*Figure 6.* NTC Automated Vehicle Program, National Transport Commission, October 2019.[20]

The SAE Standard sets out the differing levels of control required from either a human driver or an automated driving system, or combinations of both, at each specific level of automation. Levels 3 and 4 highlight the shifting responsibility for control and/oversight of the automated driving system (ADS) and is where problems [legal and otherwise] for failure to warn or failure to heed warnings situate. The most significant challenges identified by the NTC are based on unknowns around vehicle automation, including:

- The timing of deployment
- Applications that will be deployed
- The mix of technologies that automated vehicles will use
- How automated vehicles will change vehicle ownership and business models.

In order to address these challenges, the NTC aims to ensure that:

- Reforms are outcomes based, with safety as the key outcome, allowing industry to determine how best to achieve those outcomes
- Reforms are neutral as to the technologies, applications and business-models that industry develop
- Reforms are nationally consistent and internationally aligned.[21]

---

[20] National Transport Commission, Automated Vehicle Program, (October 2019) 8.

[21] National Transport Commission, Automated Vehicle Program, (October 2019) 7.

The ARRB predicts that the likely future land transport and mobility environment will be an ecosystem of operations under the current prescriptive regulatory and operational requirements, together with a variety of automated and autonomous operations in private, public and commercial transport permissioned under a safety assurance system.

Readiness for uptake of RAS-AI in Australia requires changes in technology as well as infrastructure (signals, signs and lines, pavements, lighting, connectivity, data security etc), drivers and other road users (education, licensing, including the "autonomous driving entity") and regulation and operational frameworks (driver's licenses, parking policy, overtaking, gap setting and traffic movement protocols, accident investigation, national traffic information dataset, data custodians, etc).

## Robotics standards and regulation

Frameworks to systematically and efficiently accredit RAS-AI do not yet exist, as evidenced through domains briefly explored in the previous section and emerging for new domains. Existing standards and assurance mechanisms may not be suitable for autonomous systems as they may either require human operators to be in constant control, or for robots to be kept physically separated from human operators (e.g. industrial robots).

As we have seen, technology is regulated differently depending on the sector. Currently, there are an assortment of standards and regulations emerging to govern robotics and software systems. ISO standards, for example, give assurance to industries and the community that they can trust products that have met those standards to assure citizen safety. There are a number of ISO standards applicable to robotics—see Appendix A.

The ISO group responsible for global robotics standards is ISO/TC 299.[22] At this stage, these standards cover safety requirements for industrial robots and robot systems. They are relevant to robot manufacturers, integrators and users. The standards provide good guidance for industrial robots, along with some guidance for personal care robots and medical robots. What these standards do not cover are the more diverse robotic industries and robotic software and learning systems. Software development for functional safety of industrial control systems is covered in IEC61508, IEC62061 and ISO 13849-1.

Specific assurance frameworks for different technology types are made more complex by the regulatory frameworks and legal definitions inherited from domains where they are deployed.

The challenge for the Australian robotics industry is to identify, adopt and promote trust and safety frameworks that align to a manageable assurance framework, are accepted by society, and are pragmatic for robotics industries to incorporate and abide by.

A number of initiatives and models have been proposed, or are being explored, to illustrate notions of trust and safety in autonomous systems. Some industries have *trust marks* that indicate that various security and safety requirements have been met. Trust marks have been proposed as measures to encourage more responsible technology design practice, e.g. in relation to sensitive data sharing.[23]

---

[22] ISO Technical Committee (TC) 299.

As mentioned earlier, ACcreditation-as-a-Service (ACaaS) is a "helicopter parent" approach to trust, where an API constantly pings a robot in operation and requests status updates with regards to what it is doing and why it is doing it. ACaaS could contribute to challenges identified in previous sections where a rapidly changing robot operates independently in an uncertain environment. An effective and robust communications network might allow for faster intervention if a robot did behave outside the parameters of authorised operations.

It may be that more than discrete new kinds of interventions, a complete re-evaluation of existing regulatory approaches might be needed.

## New ways of thinking about regulation

In order for the robotics industry to thrive, a regulatory response that adapts, protects, engenders community trust, while accounting for rapidly evolving industrial environments, is essential. This is an area where investment can lead to improved outcomes for the whole robotics sector, and Australia can lead the world.

The current regulation of autonomous and remotely operated vehicles requires in-depth regulatory engagement, meaning it can take much longer than usual for a manufacturer or operator to get approvals from the relevant authorities. As noted above, existing laws and regulation of land vehicles, aerial vehicles and maritime vessels rely on a human operator to be responsible for the platform, and for the platform to meet specifications intended to provide for human, platform, and environmental safety. Robotic platforms must therefore seek exemptions from some or all of current requirements to operate.

The exemption process can be long when the operator does not understand the regulatory framework and does not have prior operational data to illustrate safe operations, and the regulator does not understand the platform or the technology that will enable safe operation. This is exacerbated when an operator identifies the use of an AI-based system to mitigate an operational risk but does not explain how it works or provide evidence of its efficacy.

While traditional regulatory approaches could be characterised as prescriptive and inspection based, often reacting to problems that have occurred, more modern approaches are moving towards being risk based, responsive, and anticipatory, proactively taking opportunities to support positive innovation. Where a regulator needs to rely on exemptions, because the regulatory framework cannot keep pace with technology, it is an indicator that change is needed.

The future of adaptive regulation will mean changing the existing processes of regulation and the factors that might be considered as relevant to the regulatory process. Technologies are no

---

[23] https://www.alrc.gov.au/publication/for-your-information-australian-privacy-law-and-practice-alrc-report-108/31-cross-border-data-flows/trustmarks/.

longer based on highly deterministic systems. Regulators will need to consider society's appetite for risk and ethical expectations. To this end, Australia has adopted ethical AI principles that facilitate dialogue and design of robots—see Appendix B: Australia's AI Ethics Principles

The concept of anticipatory regulation is inclusive and collaborative, future-facing, proactive, iterative, outcomes-based and experimental—see *Figure 7*.

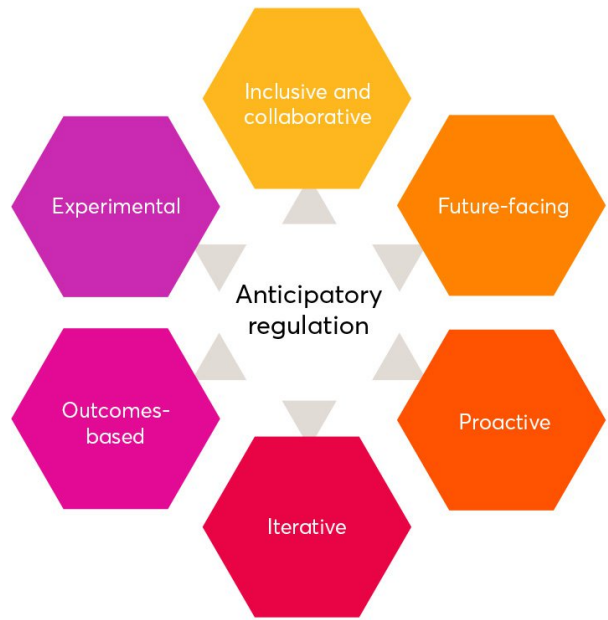

*Figure 7* Anticipatory regulation see Nesta—formerly *National Endowment for Science, Technology and the Arts*[24]

Central to the concept of anticipatory regulation is a focus on "co-design", where the regulator and industry work together to co-design standards and regulation that are fit for purpose and achieve the required outcomes. This approach focusses on the *system* and how to *influence* it. It is consistent with the concept of regulatory stewardship, whereby regulation is seen as an asset managed proactively and collaboratively to help things happen effectively. If Australian safety regulators could incorporate such approaches, and work towards becoming adaptive regulators, it may dramatically improve their ability to keep pace with technological change.

---

[24] https://www.nesta.org.uk/feature/innovation-methods/anticipatory-regulation/

Case study: Agile AI regulation by Biarri & QUT Law

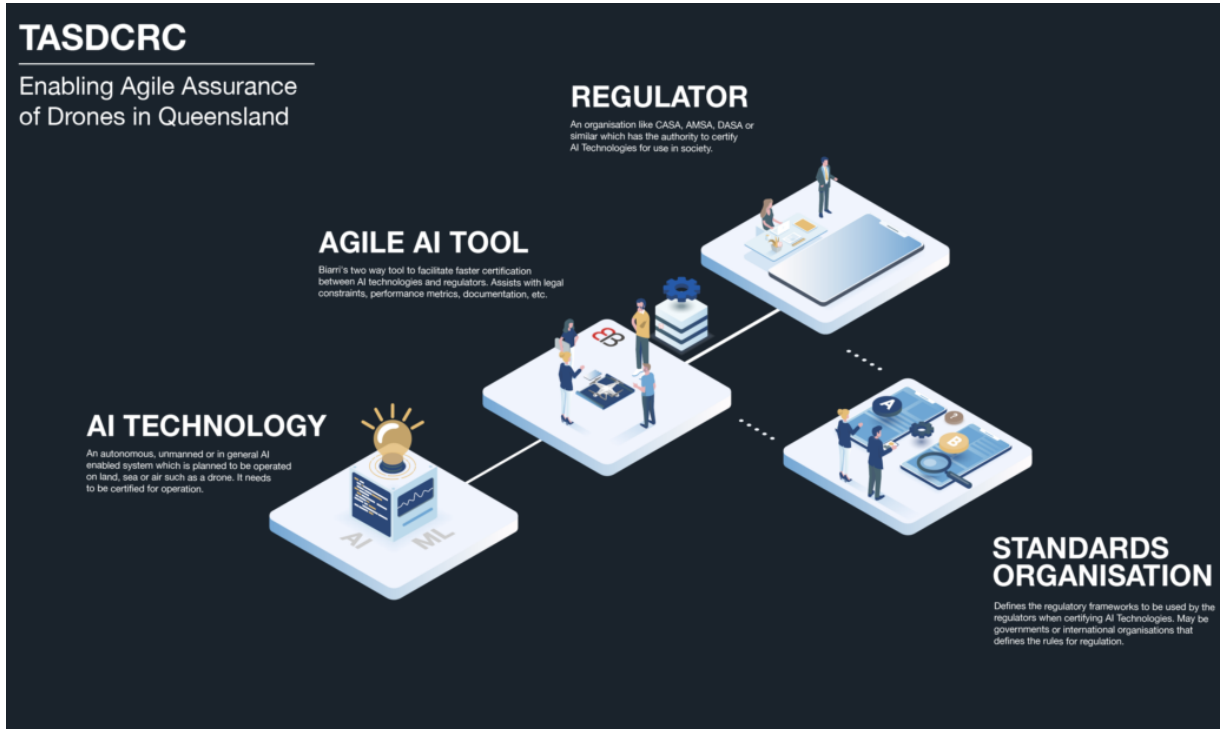

*Figure 8.* The Agile AI tool will connect the regulated and the regulator for faster assurance of AI.

Biarri provides decision support assistance and tools to customers globally, developing software applications and web-based models which are fast to run and easy to use. Biarri assists customers to find savings and improve business performance and decision making by deploying advanced mathematics, statistical and AI approaches as web-based tools using the Biarri Workbench SaaS platform.

The Workbench platform provides advanced business process optimisation and regulatory tools in a variety of industries and scenarios from the AFL to retail, and government to aviation. The platform encapsulates complex rules and mathematical engines and wraps them with simple user interfaces, configured to each use case.

Biarri's vision is to enable faster innovation cycles for AI enabled products by speeding up the development, certification and assurance of robots and autonomous systems in highly regulated industries. Currently, certification and assurance processes can take years in several industrial sectors such as aviation, maritime and more. Coupled with low sales volumes in such industries, this can significantly stifle growth and innovation. Therefore, a need exists for a set of software tools which accelerate and simplify entire certification processes – via leveraging digital platforms which are tailored to the requirements of regulating AI enabled products. These types of software tools have revolutionised many other industries and so the timing is right to disrupt the regulation industry via digital transformation. By enabling more innovation, such tools can catalyse the next wave of autonomous systems to solve large scale global challenges.

The emergence of AI systems suggests two challenges for the future of regulation. The first is the immediate context around regulating the application of novel AI technologies. This, on the surface, is about identification of desirable outcomes and ends, and the deployment of an appropriate mix of regulatory strategies directed to those outcomes and ends. However, the emergence of AI is more significant. AI enabled systems have the potential to be an important regulatory strategy in their own right. This second challenge has been the focus in recent computational law literature examining how AI enabled systems can be developed as regulatory instruments.

QUT Law sees this project as highly innovative as it is located exactly at the nexus between the regulation of AI and AI as regulation. This extends to a variety of fields. First it connects to recent work on smart and agile regulation; through examining how to build automated digital systems that allow confidence and trust between regulators and regulatees; in this project – specifically between the AUX innovators and transport safety regulators. Second, it engages with the core project of computational law on the theory and practice of translating established legal forms into digital platforms through the process of digitalising Australian maritime safety regulations. Third, it will explore how sandbox design and concepts familiar in the "fintech" space, could be used to enhance confidence and trust between AI transport innovators and Australian regulators.

## Designing Democratically Legitimate AI Systems

Trust, safety, and assurance are all values that depend on other values as well as reliable critical infrastructure[25]. People can trust a system when they believe that it will act in conformity with shared values as well as being technically robust. A system is safe when people can rely on it not to harm them and others or damage things that they care about. A system provides assurance when it provides robust evidence of its safety and trustworthiness. To design robotic systems with Trust, Safety, and Assurance in mind, there is still basic research to design such values into automated systems more generally.

But which values? How can morally justified robotic systems be designed, given how much moral disagreement there is in society?

### Case study: The Humanising Machine Intelligence project, ANU

The Humanising Machine Intelligence project, at ANU, was founded on the belief that the moral disagreement faced by designers of AI and robotic systems is no different from moral disagreement elsewhere in society. And the only viable solution ever developed for accommodating moral disagreement without mass oppression is democracy. Of course, democracies must respect fundamental human rights, cultural group rights, indigenous land claims, and much more. But the key point is that the process of selecting the values designed into AI systems should be the same as our process for resolving other evaluative conflicts—we need to *adapt* democratic institutions to play this role, as we must adapt them continuously for all social and technological changes, but democratic self-determination should be our lodestar. Once we settle on how to decide which values to deploy, we face the pressing question of *how* to design them into feasible AI and robotic systems. To make any real progress, we obviously need to draw on different domains of knowledge, and different experiences. No single methodology has a monopoly on truth, and the greatest progress will be made by working in concert across disciplinary boundaries. The HMI project unites world-leading experts and rising stars in computer science, philosophy, political science, law and sociology. We have framed our research questions collaboratively, working also with partners in government, industry and civil society. And we have helped build a diverse community of researchers working on Data, AI and Society in Australia, and around the world. We use all of our disciplines to give a comprehensive picture of each problem we address: we start by using our more empirical, positivist disciplines to give an account of the opportunities and risks associated with particular applications of data and AI in the world; we then rely on our theoretical disciplines to provide the moral diagnosis of that empirical research, and chart a course to where we need to be. On those foundations, we deploy our technical disciplines to build democratically legitimate data and AI systems.

---

[25] See arguments pertaining to networks, data, systems in The Department of Home Affairs Consultation Paper 'Protecting Critical Infrastructure and Systems of National Significance' (Aug 2020) https://www.homeaffairs.gov.au/reports-and-publications/submissions-and-discussion-papers/protecting-critical-infrastructure-systems

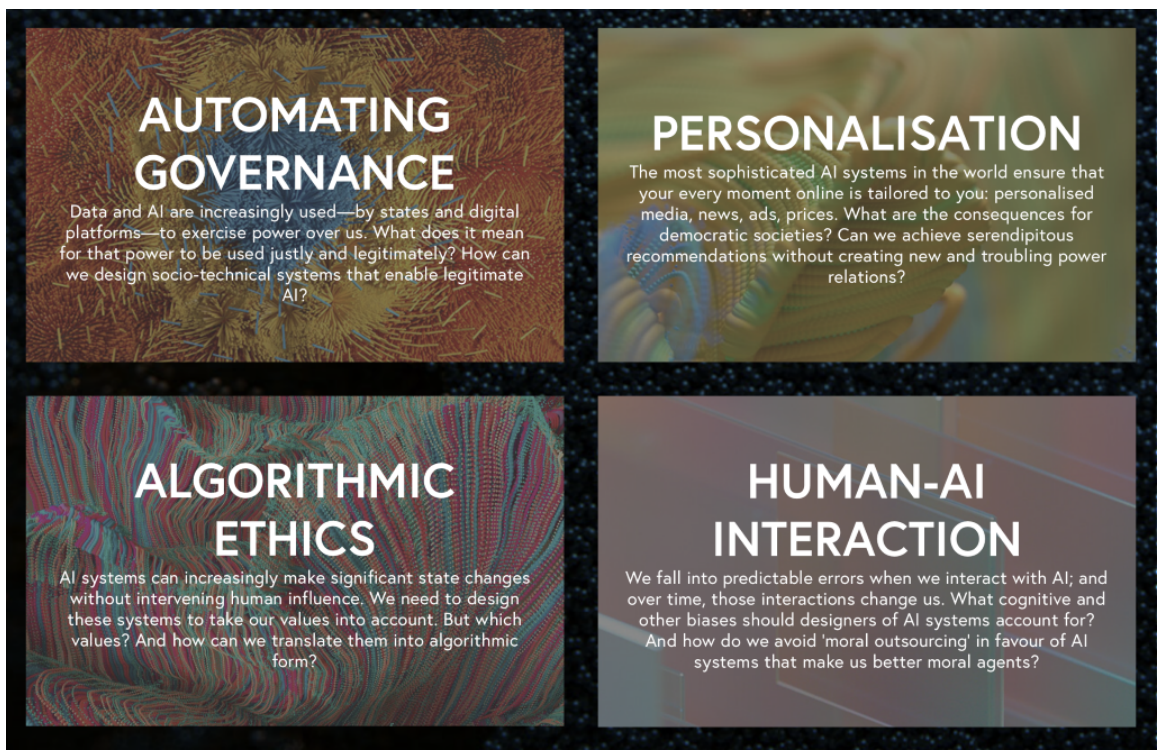

*Figure 9.* Humanising Machine Intelligence project fundamental research areas include automating governance, personalisation, algorithmic ethics and the ethics of human-AI interaction.[26]

---

[26] https://hmi.anu.edu.au/research.

We focus on four key research themes: Automating governance, personalisation, algorithmic ethics and human-AI interaction—see Figure 6. Of those, two are especially relevant for the design of robotic systems that are trustworthy, safe, and assured. If we are going to design our values into automated systems (robotic or otherwise), we have to represent those values in a language that computers can act on.  How can we faithfully represent moral reasons within computer algorithms without unintended (and potentially serious) consequences? That is the task of our "**Algorithmic Ethics**" project. But as Rousseau wrote, "our goal when designing the ideal society should be to imagine the laws as they might be, but to take people as they are". If we design algorithmic systems that assume that the humans they work with are as rational as they are, then we're set for disaster. We need to design robotic systems that recognise the predictable cognitive biases into which their "human in the loop" will inevitably fall, as well as the ways in which human-machine teaming will change us, as people. Understanding and solving for those challenges is the task of our **"Ethics of Human-AI Interaction"** project.

## Designing automated systems holistically for trust and safety

Complex learning systems today carry a range of hardware and software safety issues to consider, as acknowledged in 'Emerging Issues in Assuring Autonomous Robots'. Considering direct harms that could arise from complex learning systems, as well as indirect harms or unexpected consequences, necessitates new, creative ways of thinking about sustaining safety and trust.

The 3Ai, an innovation institute based at the Australian National University and founded by Distinguished Professor Genevieve Bell, is exploring ways in which cybernetics, the transdisciplinary framework that influenced disciplines as diverse as systems engineering, artificial intelligence and management science in the mid 20th century, could be extended or revised to account for complex learning systems today. While cybernetics has itself undergone several transformations, its central preoccupation with feedback loops - both within technical systems, and between systems and their physical, human and bureaucratic environments – provide an interconnected, nuanced way of exploring complex concepts like "trust" and "safety".

Establishing and communicating that a system is "safe" invokes a range of interventions and practices, touched on across this chapter – laws, regulations and standards; audit, verification and validation processes; forms of independent monitoring and review; workplace training and process; and broader safety culture. Cybernetics, and the range of disciplines 3Ai staff come from, including systems engineering, nuclear physics, computer science, medical anthropology, journalism and data science, have enabled 3Ai to explore the connections between these practices. 3Ai combines science, analytics, history and art to explore complex systems.

Case study: Using cybernetics to explore trust and safety of AI on the Great Barrier Reef

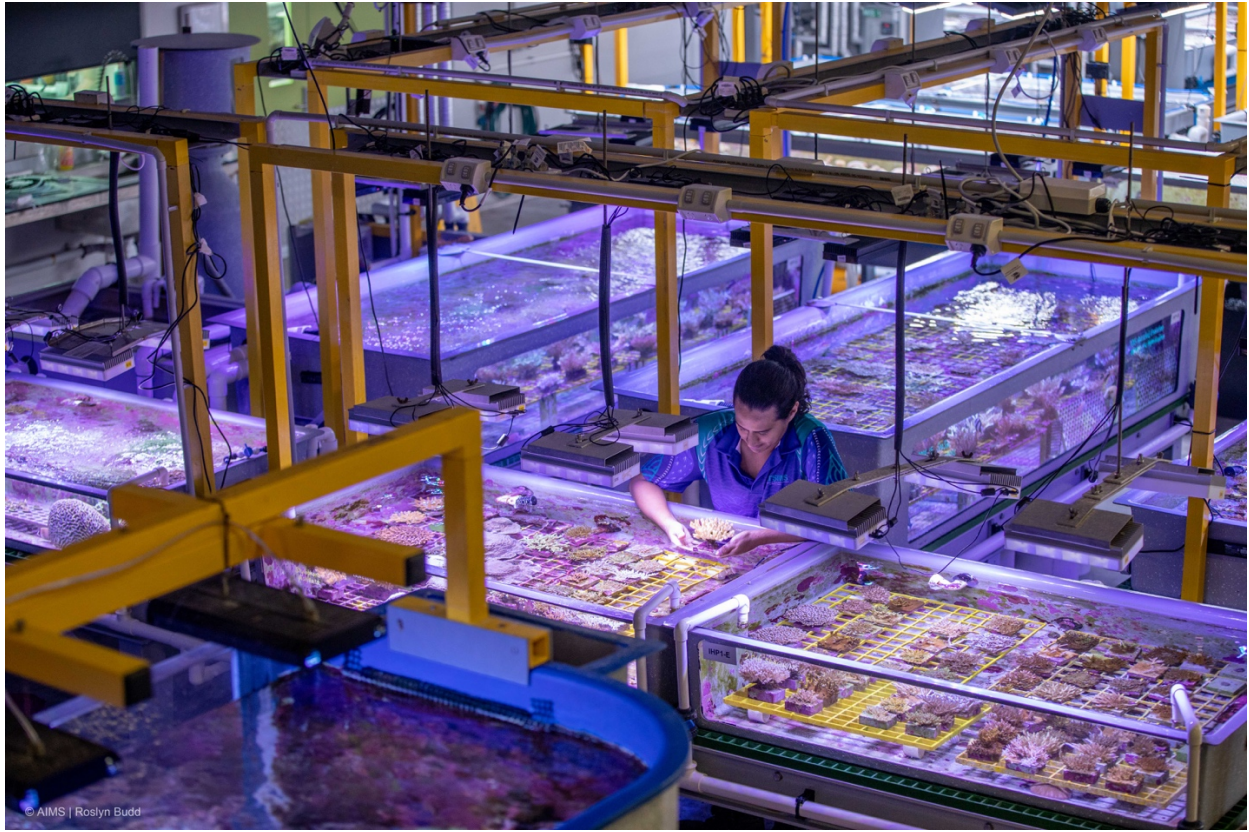

*Figure 10*. SeaSim at AIMS uses models of ocean conditions to automatically simulate conditions in tanks, in order to observe coral behaviour. Credit Roslyn Budd

With the Australian Institute of Marine Science (AIMS), 3Ai explored the design and implementation of automated underwater systems monitoring the health of the Great Barrier Reef. 3Ai explored the influences behind this transition to more autonomous technology in addition to the impact on the practices and expectations of AIMS as an organisation. The Great Barrier Reef Marine Park is one of the world's largest marine protected areas, and the proper management and protection of this world heritage site, across an area the size of Italy in conditions that are often challenging, requires AIMS to continue to improve its monitoring and data collection practices. AIMS is exploring using above water and under water cyber-physical systems to monitor the reef environment without increasing human involvement.

3Ai combined interviews with AIMS scientists and technicians with origin stories, organisational discourse analysis and critical systems heuristics to explore how the design and deployment of automated systems was being perceived across AIMS. Using these methods, 3Ai and AIMS have been exploring a range of effects of this transition, including relationships between trust in data collected by automated systems, as compared with data traditionally collected by humans. Direct, lived [human] experience working on the reef, and the ability to draw on that experience and common sense to identify mistakes, were highlighted as factors influencing trust in human

divers collecting data. 3Ai observed that the mere fact of a human diver's lived experience - their humanity - engendered a degree of trust in diver collected data that resulted in an inherently higher level of confidence and trust in comparison to the automated equivalent. This work is ongoing.

## Conclusion

Robotics in Australia have a long history of conforming with safety standards and risk managed practises. This chapter articulates the current state of trust and safety in robotics including society's expectations, safety management systems and system safety as well as emerging issues and methods for ensuring safety in increasingly autonomous robotics.

The future of trust and safety will combine standards with iterative, adaptive and responsive regulatory and assurance methods for diverse applications of RAS-AI. Robotics will need novel technical and social approaches to achieve assurance, particularly for game-changing innovations. The complexity of RAS-AI calls for transdisciplinary collaboration across technical, scientific, humanities disciplines as well as outreach to stakeholders of human-RAS-AI interaction.

Novel technical approaches for those assurances are required. The ability for users to easily update algorithms and software, which alters the performance of a system, implies that traditional machine assurance performed prior to deployment or sale, will no longer be viable. Moreover, the high frequency of updates implies that traditional certification that requires substantial time will no longer be practical. To alleviate these difficulties, automation of assurance will likely be needed; something like 'ASsurance-as-a-Service' (ASaaS), where APIs constantly ping RAS-AI to ensure abidance with various rules, frameworks, and behavioural expectations. There are exceptions to this, such as in contested or communications denied environments, or in underground or undersea mining; and these systems need their own risk assessments and limitations imposed. Indeed, self-monitors are already operating within some systems.

The assurance process will require stakeholders to possess sufficient technological and computational skills, knowledge and abilities. To ensure safe operation of future robotics systems, Australia needs to invest in RAS-AI assurance research, stakeholder engagement and continued development and refinement of robust frameworks, methods, guidelines and policy in order to educate and prepare its technology developers, certifiers, and general population.

# Appendices

**APPENDIX A: STANDARD AND/OR PROJECT UNDER THE DIRECT RESPONSIBILITY OF ISO/TC 299 SECRETARIAT (Standardization in the field of robotics, excluding toys and military applications)[27]**

ISO/WD 5124 Robotics — Services provided by service robots — Safety management systems requirements

ISO 8373:2012 Robots and robotic devices — Vocabulary

ISO/DIS 8373 Robotics — Vocabulary

ISO 9283:1998 Manipulating industrial robots — Performance criteria and related test methods

ISO 9409-1:2004 Manipulating industrial robots — Mechanical interfaces — Part 1: Plates

ISO 9409-2:2002 Manipulating industrial robots — Mechanical interfaces — Part 2: Shafts

ISO 9787:2013 Robots and robotic devices — Coordinate systems and motion nomenclatures

ISO 9946:1999 Manipulating industrial robots — Presentation of characteristics

ISO 10218-1:2011 Robots and robotic devices — Safety requirements for industrial robots — Part 1: Robots

ISO/DIS 10218-1 Robotics — Safety requirements for robot systems in an industrial environment — Part 1: Robots

ISO 10218-2:2011 Robots and robotic devices — Safety requirements for industrial robots — Part 2: Robot systems and integration

ISO/CD 10218-2 Robotics — Safety requirements for robotics in an industrial environment — Part 2: Robot systems and integration

ISO 11593:1996 Manipulating industrial robots — Automatic end effector exchange systems — Vocabulary and presentation of characteristics

ISO/DIS 11593 Robots for industrial environments — Automatic end effector exchange systems — Vocabulary and presentation of characteristics

ISO/TR 13309:1995 Manipulating industrial robots — Informative guide on test equipment and metrology methods of operation for robot performance evaluation in accordance with ISO 9283

---

[27] https://www.iso.org/committee/5915511/x/catalogue/

ISO 13482:2014 Robots and robotic devices — Safety requirements for personal care robots

ISO 14539:2000 Manipulating industrial robots — Object handling with grasp-type grippers — Vocabulary and presentation of characteristics

ISO/TS 15066:2016 Robots and robotic devices — Collaborative robots

ISO 18646-1:2016 Robotics — Performance criteria and related test methods for service robots — Part 1: Locomotion for wheeled robots

ISO 18646-2:2019 Robotics — Performance criteria and related test methods for service robots — Part 2: Navigation

ISO/DIS 18646-3 Robotics — Performance criteria and related test methods for service robots — Part 3: Manipulation

ISO/DIS 18646-4 Robotics — Performance criteria and related test methods for service robots — Part 4: Lower-back support robots

ISO 19649:2017 Mobile robots — Vocabulary

ISO/TR 20218-1:2018 Robotics — Safety design for industrial robot systems — Part 1: End-effectors

ISO/TR 20218-2:2017 Robotics — Safety design for industrial robot systems — Part 2: Manual load/unload stations

ISO/FDIS 22166-1 Robotics — Modularity for service robots — Part 1: General requirements

ISO/TR 23482-1:2020 Robotics — Application of ISO 13482 — Part 1: Safety-related test methods

ISO/TR 23482-2:2019 Robotics — Application of ISO 13482 — Part 2: Application guidelines

IEC/TR 60601-4-1:2017 Medical electrical equipment — Part 4-1: Guidance and interpretation — Medical electrical equipment and medical electrical systems employing a degree of autonomy

IEC 80601-2-77:2019 Medical electrical equipment — Part 2-77: Particular requirements for the basic safety and essential performance of robotically assisted surgical equipment

IEC 80601-2-78:2019 Medical electrical equipment — Part 2-78: Particular requirements for basic safety and essential performance of medical robots for rehabilitation, assessment, compensation or alleviation

**APPENDIX B: AUSTRALIA'S AI ETHICS PRINCIPLES[28]**

1. Human, social and environmental wellbeing: Throughout their lifecycle, AI systems should benefit individuals, society and the environment.
2. Human-centred values: Throughout their lifecycle, AI systems should respect human rights, diversity, and the autonomy of individuals.
3. Fairness: Throughout their lifecycle, AI systems should be inclusive and accessible, and should not involve or result in unfair discrimination against individuals, communities or groups.
4. Privacy protection and security: Throughout their lifecycle, AI systems should respect and uphold privacy rights and data protection, and ensure the security of data.
5. Reliability and safety: Throughout their lifecycle, AI systems should reliably operate in accordance with their intended purpose.
6. Transparency and explainability: There should be transparency and responsible disclosure to ensure people know when they are being significantly impacted by an AI system, and can find out when an AI system is engaging with them.
7. Contestability: When an AI system significantly impacts a person, community, group or environment, there should be a timely process to allow people to challenge the use or output of the AI system.
8. Accountability: Those responsible for the different phases of the AI system lifecycle should be identifiable and accountable for the outcomes of the AI systems, and human oversight of AI systems should be enabled.

---

[28] https://www.industry.gov.au/data-and-publications/building-australias-artificial-intelligence-capability/ai-ethics-framework/ai-ethics-principles